\pdfoutput=1

\documentclass[11pt]{article}

\usepackage{EMNLP2022}

\usepackage{times}
\usepackage{latexsym}

\usepackage{bm}
\usepackage{mathtools}
\usepackage{float}
\usepackage[ruled,vlined,linesnumbered]{algorithm2e}
\usepackage{algpseudocode}
\usepackage{leftidx}
\usepackage{url}
\usepackage{tabularx, multirow, multicol}
\usepackage{color, colortbl}
\usepackage{soul}
\usepackage{caption}
\usepackage{subcaption}
\usepackage{float}
\usepackage{balance}
\usepackage{setspace}
\usepackage{bbm}
\usepackage{pifont}
\usepackage{microtype}
\usepackage[normalem]{ulem}
\usepackage{wasysym}

\usepackage{amsmath}
\usepackage{amsfonts}
\usepackage{booktabs}
\usepackage{graphicx}
\usepackage{dsfont}
\usepackage{xspace}
\usepackage{hyperref}

\definecolor{Gray}{gray}{0.95}

\newcolumntype{B}{>{\raggedright\arraybackslash}m{0.7\linewidth}}
\newcolumntype{P}{>{\centering\arraybackslash}m{0.16\linewidth}}
\newcolumntype{Q}{>{\raggedright\arraybackslash}m{0.6\linewidth}}
\newcolumntype{R}{>{\raggedright\arraybackslash}m{0.71\linewidth}}
\newcolumntype{T}{>{\raggedright\arraybackslash}m{0.61\linewidth}}
\newcolumntype{U}{>{\centering\arraybackslash}m{0.23\linewidth}}
\newcolumntype{V}{>{\centering\arraybackslash}m{0.18\linewidth}}

\newcolumntype{S}{>{\raggedright\arraybackslash}m{0.5\linewidth}}
\newcolumntype{W}{>{\raggedright\arraybackslash}m{0.55\linewidth}}

\usepackage[T1]{fontenc}

\usepackage[utf8]{inputenc}

\usepackage{microtype}

\usepackage{inconsolata}

\newenvironment{talign*}
 {\csname align*\endcsname}
 {\endalign}

\newcommand{\glove}{GloVe\xspace}

\newcommand{\textrank}{TextRank\xspace}
\newcommand{\topicrank}{TopicRank\xspace}
\newcommand{\topickg}{TopicKG\xspace}
\newcommand{\bertkg}{BERT-KG\xspace}
\newcommand{\berttkg}{BERT-TKG\xspace}

\newcommand{\hlda}{hLDA\xspace}

\newcommand{\taxogen}{TaxoGen\xspace}
\newcommand{\taxocom}{TaxoCom\xspace}
\newcommand{\corel}{CoRel\xspace}

\newcommand{\proposedab}{TopicExpan\textsuperscript{-sr}\xspace}
\newcommand{\proposedfull}{TopicExpan\textsuperscript{+sr}\xspace}
\newcommand{\proposed}{TopicExpan\xspace}

\newcommand{\amazon}{Amazon\xspace}
\newcommand{\dbpedia}{DBPedia\xspace}

\newcommand{\cvec}[1]{\bm{c}_{#1}}
\newcommand{\svec}[1]{\bm{s}_{#1}}
\newcommand{\dvec}[1]{\bm{d}_{#1}}

\newcommand{\vvec}[1]{\bm{v}_{#1}}

\newcommand{\hvec}[1]{\bm{h}_{#1}}

\newcommand{\doc}[1]{d_{#1}}
\newcommand{\topic}[1]{c_{#1}}
\newcommand{\phrase}[1]{p_{#1}}
\newcommand{\token}[1]{v_{#1}}
\newcommand{\topicphs}[1]{\mathcal{P}_{#1}}

\newcommand{\vtopic}[1]{c^*_{#1}}
\newcommand{\vphrase}[1]{p^*_{#1}}

\newcommand{\simloss}{\mathcal{L}_{sim}}
\newcommand{\genloss}{\mathcal{L}_{gen}}

\newcommand{\vocaset}{\mathcal{V}}
\newcommand{\docuset}{\mathcal{D}}
\newcommand{\cateset}{\mathcal{C}}
\newcommand{\edgeset}{\mathcal{R}}
\newcommand{\taxo}{\mathcal{T}}
\newcommand{\subtaxo}[1]{\mathcal{T}_{#1}}

\newcommand{\wrong}[1]{\st{#1}}
\newcommand{\unique}[1]{\ul{#1}}
\newcommand{\reducetxt}[1]{\textls[-5]{#1}}
\newcommand{\reduce}[1]{\textls[-50]{#1}}

\newcommand{\smallsection}[1]{{\vspace{0.03in} \noindent \bf {#1.\hspace{5pt}}}}

\title{\reducetxt{Topic Taxonomy Expansion via Hierarchy-Aware Topic Phrase Generation}}

\author{Dongha Lee\textsuperscript{1}\hspace{-3pt}, Jiaming Shen\textsuperscript{2}\hspace{-3pt}, Seonghyeon Lee\textsuperscript{3}\hspace{-3pt}, Susik Yoon\textsuperscript{1}\hspace{-3pt}, Hwanjo Yu\textsuperscript{3}\thanks{\ \ Corresponding author}, and Jiawei Han\textsuperscript{1} \\
  \textsuperscript{1}University of Illinois at Urbana-Champaign (UIUC), Urbana, IL, United States \\
  \textsuperscript{2}Google Research, New York, NY, United States \\
  \textsuperscript{3}Pohang University of Science and Technology (POSTECH), Pohang, Republic of Korea \\
  {\{donghal,susik,hanj\}@illinois.edu, jmshen@google.com, \{sh0416,hwanjoyu\}@postech.ac.kr}\\}

\begin{document}
\maketitle
\begin{abstract}
\reducetxt{Topic taxonomies display hierarchical topic structures of a text corpus and provide topical knowledge to enhance various NLP applications.
To dynamically incorporate new topic information, several recent studies have tried to expand (or complete) a topic taxonomy by inserting emerging topics identified in a set of new documents.
However, existing methods focus only on frequent terms in documents and the local topic-subtopic relations in a taxonomy, which leads to limited topic term coverage and fails to model the global topic hierarchy.
In this work, we propose a novel framework for topic taxonomy expansion, named \proposed, which directly generates topic-related terms belonging to new topics.
Specifically, \proposed leverages the hierarchical relation structure surrounding a new topic and the textual content of an input document for topic term generation.
This approach encourages newly-inserted topics to further cover important but less frequent terms as well as to keep their relation consistency within the taxonomy.
Experimental results on two real-world text corpora show that \proposed significantly outperforms other baseline methods in terms of the quality of output taxonomies.}
\end{abstract}

\section{Introduction}
\label{sec:intro}
Topic taxonomy is a tree-structured representation of hierarchical relationship among multiple topics found in a text corpus~\cite{zhang2018taxogen, shang2020nettaxo, meng2020hierarchical}. 
Each topic node is defined by a set of semantically coherent terms related to a specific topic (i.e., topic term cluster), and each edge implies the ``general-specific'' relation between two topics (i.e., topic-subtopic).
With the knowledge of hierarchical topic structures, topic taxonomies have been successfully utilized in many text mining applications, such as text summarization~\cite{petinot2011hierarchical, bairi2015summarization} and categorization~\cite{meng2019weakly, shen2021taxoclass}.

\begin{figure}[t]
    \centering
    \includegraphics[width=\linewidth]{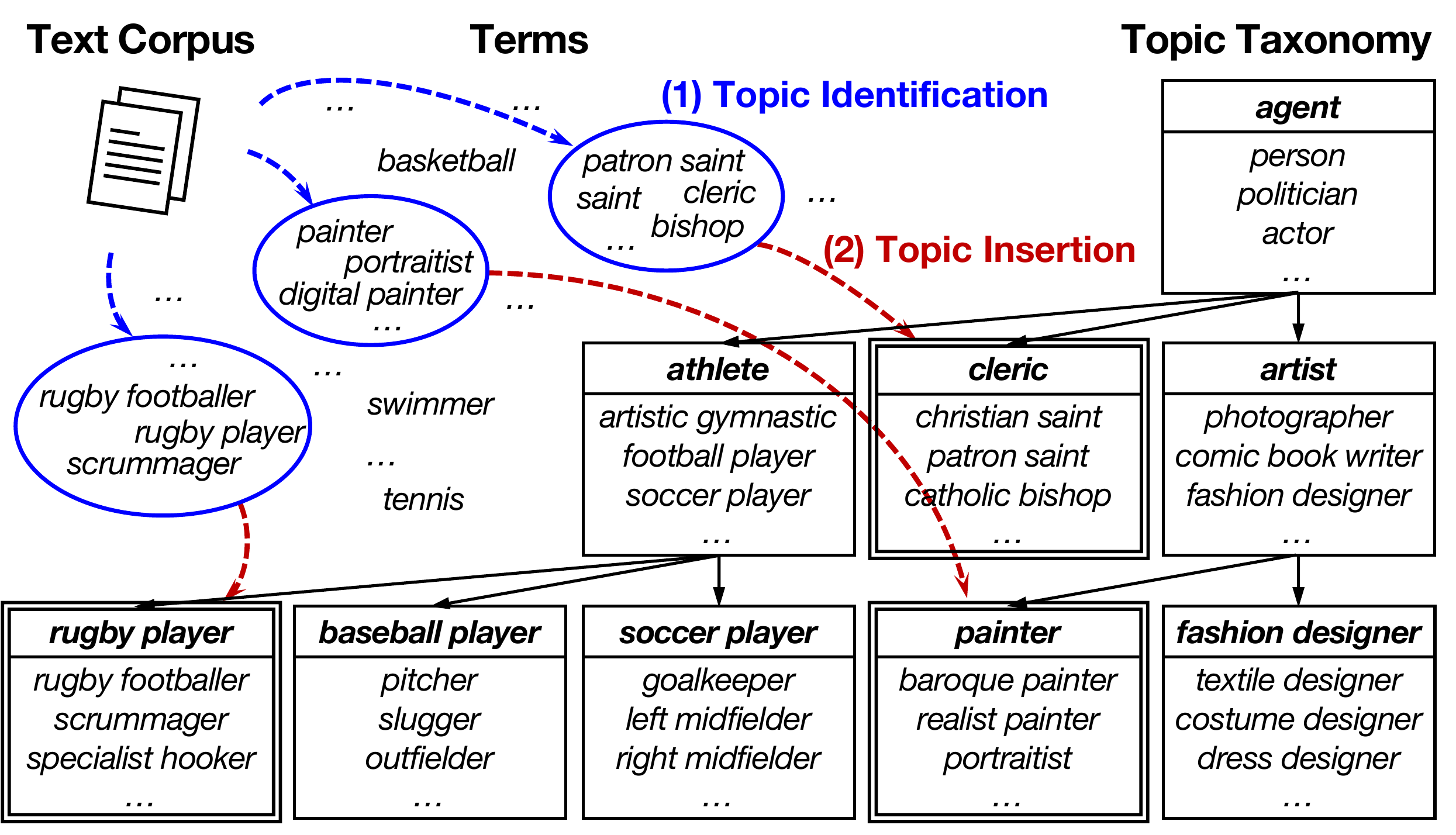}
    \caption{An example of topic taxonomy expansion. The known (i.e., existing) topics and novel topics are in single-line and double-line boxes, respectively.}
    \label{fig:problem}
\end{figure}

Recently, automated expansion (or completion) of an existing topic taxonomy has been studied \cite{huang2020corel, lee2022taxocom}, which helps people to incrementally manage the topic knowledge within fast-growing document collections.
This task has two technical challenges:
(1) identifying new topics by collecting topic-related terms that have novel semantics, and (2) inserting the new topics at the right position in the hierarchy.
In Figure~\ref{fig:problem}, for example, a new topic node \textit{painter} that consists of its topic-related terms [\textit{baroque painter}, \textit{realist painter}, \textit{portraitist}, ...] is inserted at the child position (i.e., subtopic) of the existing topic node \textit{artist}, without breaking the consistency of topic relations with the neighbor nodes.

The existing methods for topic taxonomy expansion, however, suffer from two major limitations:
(1) \textit{Limited term coverage} --
They identify new topics from a set of candidate terms, while relying on entity extraction tools~\cite{zeng2020tri} or phrase mining techniques~\cite{liu2015mining,shang2018automated,gu2021ucphrase} to obtain the high-frequency candidate terms in a corpus.
Such extraction techniques will miss a lot of topic-related terms that have low frequency, and thus lead to an incomplete set of candidate terms~\cite{zeng2021enhancing}.
(2) \textit{Inconsistent topic relation} --
As they insert new topics by considering only the first-order relation between two topics (i.e., a topic and its subtopic), the newly-inserted topics are likely to have inconsistent relations with other existing topics.
The expansion strategy based on the first-order topic relation is inadequate to capture the holistic structure information of the existing topic taxonomy.

As a solution to both challenges, we present \proposed, a new framework that expands the topic taxonomy via \textit{hierarchy-aware topic term generation}.
The key idea is to directly generate topic-related terms from documents by taking the topic hierarchy into consideration.
From the perspective of term coverage, this generation-based approach can identify more multi-word terms even if they have low frequency in the given corpus~\cite{zeng2021enhancing}, compared to the extraction-based approach only working on the extracted candidate terms that frequently appear in the corpus.
To combat the challenge of relation inconsistency, we utilize graph neural networks (GNNs) to encode the relation structure surrounding each topic~\cite{kipf2017semi, shen2021taxoclass} and generate topic-related terms conditioned on these relation structure encodings.
This allows us to accurately capture a hierarchical structure beyond the first-order relation between two topics.

To be specific, \proposed consists of \textit{the training step} and \textit{the expansion step}.
The training step is for optimizing a neural model that topic-conditionally generates a term from an input document.
Technically, for topic-conditional term generation, the model utilizes the relation structure of a topic node as well as the textual content of an input document.
The expansion step is for discovering novel topics and inserting them into the topic taxonomy.
To this end, \proposed places a \textit{virtual} topic node underneath each existing topic node, and then it generates the topic terms conditioned on the virtual topic by utilizing the trained model.
In the end, it performs clustering on the generated terms to identify multiple novel topics, which are inserted at the position of the virtual topic node.


\smallsection{Contributions}
The main contributions of this paper can be summarized as follows:
(1) We propose a novel framework for topic taxonomy expansion, which tackles the challenges in topic term coverage and topic relation consistency via hierarchy-aware topic term generation.
(2) We present a neural model to generate a topic-related term from an input document \textit{topic-conditionally} by capturing the hierarchical relation structure surrounding each topic based on GNNs.
(3) Our comprehensive evaluation on two real-world datasets demonstrates that output taxonomies of \proposed show better relation consistency as well as term coverage, compared to that of other baseline methods.
    

\section{Related Work}
\label{sec:related}

\smallsection{Topic Taxonomy Construction}
\label{subsec:topictaxo}
To build a topic taxonomy of a given corpus from scratch, the state-of-the-art methods have focused on finding out discriminative term clusters in a hierarchical manner~\cite{zhang2018taxogen, meng2020hierarchical, shang2020nettaxo}. 
Several recent studies have started to enrich and expand an existing topic taxonomy by discovering novel topics from a corpus and inserting them into the taxonomy~\cite{huang2020corel, lee2022taxocom}.
They leverage the initial topic taxonomy as supervision for learning the hierarchical relation among topics.
To be specific, they discover new subtopics that should be inserted at the child of each topic,
by using a relation classifier trained on (parent, child) topic pairs~\cite{huang2020corel} or performing novel subtopic clustering~\cite{lee2022taxocom}. 
However, all the methods rely on candidate terms extracted from a corpus and also consider only the first-order relation between two topics, which degrades the term coverage and relation consistency of output topic taxonomies.

\smallsection{GNN-based Taxonomy Expansion}
\label{subsec:taxoexpan}
Recently, there have been several attempts to employ GNNs for expanding a given entity taxonomy~\cite{mao2020octet,shen2020taxoexpan,zeng2021enhancing}.
Their goal is to figure out the correct position where a new entity should be inserted, by capturing structural information of the taxonomy based on GNNs.
They mainly focus on an entity taxonomy that shows the hierarchical semantic relation among fine-grained entities (or terms), requiring plenty of nodes and edges in a given taxonomy to effectively learn the inter-entity relation.
In contrast, a topic taxonomy represents coarse-grained topics (or high-level concepts) that encode \textit{discriminative term meanings} as well as \textit{term co-occurrences} in documents (Figure~\ref{fig:problem}), which allows its node to correspond to a topic class of documents.
That~is, it is not straightforward to apply such methods to a topic taxonomy with much fewer nodes and edges, and thus how to enrich a topic taxonomy~with GNNs remains an important research question.

\begin{figure*}[t]
    \centering
    \includegraphics[width=\linewidth]{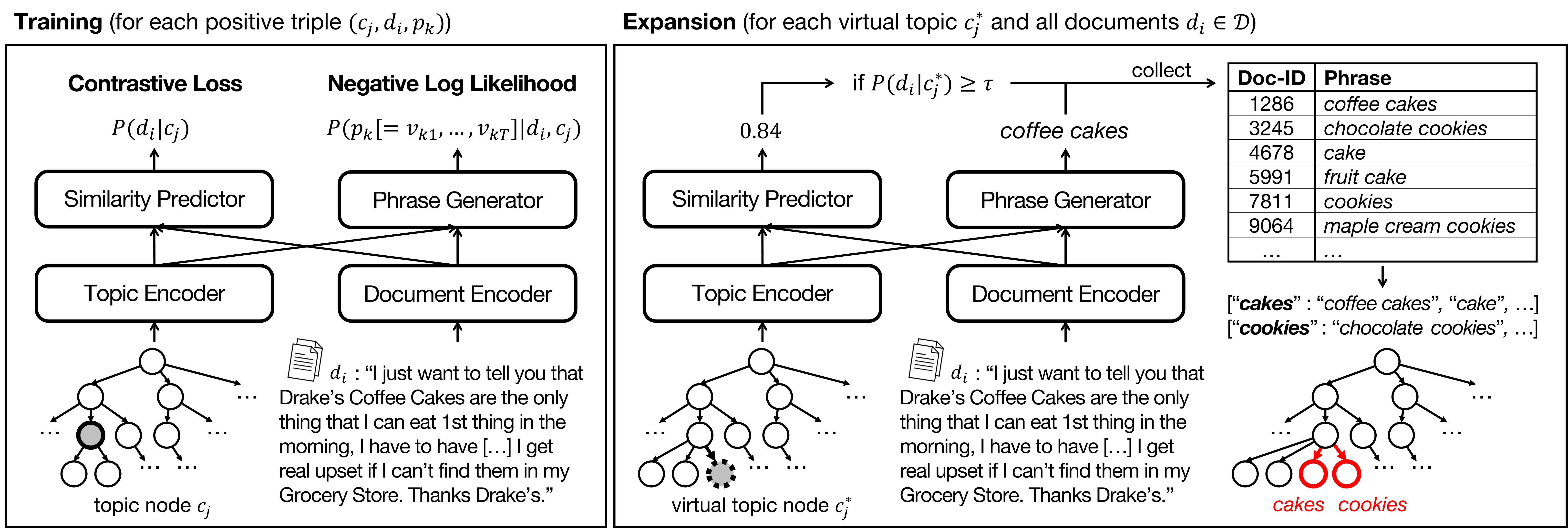}
    \caption{The overall process of \proposed. 
    (Left) It trains a unified model via multi-task learning of topic-document similarity prediction and topic-conditional phrase generation.
    (Right) It selectively collects the phrases conditionally-generated for a virtual topic node, and then it identifies multiple novel topics from phrase clusters. 
    }
    \label{fig:framework}
\end{figure*}

\smallsection{Keyphrase Generation}
\label{subsec:kpg}
The task of keyphrase prediction aims to find condensed terms that concisely summarize the primary information of an input document~\cite{liu2020keyphrase}.
The state-of-the-art approach to this problem is modeling it as the text generation task, which sequentially generates word tokens of a keyphrase~\cite{meng2017deep, zhou2021topic}.
They adopt neural architectures as a text encoder and decoder, such as an RNN/GRU~\cite{meng2017deep, wang2019topic} and a transformer~\cite{zhou2021topic}. 
Furthermore, several methods have incorporated a neural topic model into the generation process~\cite{wang2019topic, zhou2021topic} to fully utilize the topic information extracted in an unsupervised way.
Despite their effectiveness, none of them has focused on \textit{topic-conditional} generation of keyphrases from a document, as well as \textit{hierarchical} modeling of topic relations.



\section{Problem Formulation}
\label{sec:problem}

\smallsection{Notations}
A \textit{topic taxonomy} $\taxo=(\cateset, \edgeset)$ is a tree structure about topics, where each node ($\in\cateset$) represents a single conceptual topic and each edge ($\in\edgeset$) implies the hierarchical relation between a topic and its subtopic.
A topic node $\topic{j}\in\cateset$ is described by the set of topic-related terms, denoted by $\topicphs{j}$ (i.e., term cluster for the topic $\topic{j}$), where the most representative term (i.e., center term) serves as the topic name.
Each document $\doc{i}=[\token{i1}, \ldots, \token{iL}]$ and each term\footnote{Note that our proposed approach considers each term as a sequence of word tokens, and this enables us to handle a much larger number of multi-word terms, compared to the case of using a precomputed set of candidate terms.}
$\phrase{k}=[\token{k1}, \ldots, \token{kT}]$ in a given corpus $\docuset$ is the sequence of $L$ and $T$ word tokens in the vocabulary set $v\in\vocaset$, respectively.
Here, each term is regarded as a phrase that consists of one or more word tokens, so the terms ``phrase'' and ``term'' are used interchangeably in this paper.

\smallsection{Problem Definition}
Given a text corpus $\docuset$ and an initial topic taxonomy $\taxo$, the task of topic taxonomy expansion aims to discover novel topics by collecting the topic-related terms from $\docuset$ and insert them at the right position in $\taxo$ (Figure~\ref{fig:problem}).


\section{\proposed: Proposed Framework}
\label{sec:method}

\subsection{Overview}
\label{subsec:overview}
\proposed consists of (1) \textit{the training step} that trains a neural model for generating phrases topic-conditionally from documents (Figure~\ref{fig:framework} Left)
and (2) \textit{the expansion step} that identifies novel topics for each new position in the taxonomy by using the trained model (Figure~\ref{fig:framework} Right).
The detailed algorithm is described in Section~\ref{subsec:pseudocode}.


\smallsection{Training Step}
\label{subsubsec:training}
\proposed optimizes parameters of its neural model to maximize the total likelihood of the initial taxonomy $\taxo$ given the corpus $\docuset$.
\begin{equation}
\small
\label{eq:likelihood}
    \begin{split}
        P(\taxo;\docuset) &= \prod_{\topic{j}\in\cateset} \prod_{\phrase{k}\in\topicphs{j}} P(\phrase{k}|\topic{j};\docuset) \\
        &= \prod_{\topic{j}\in\cateset} \prod_{\phrase{k}\in\topicphs{j}} \sum_{\doc{i}\in\docuset} P(\phrase{k}, \doc{i} |\topic{j}) \\
        &\approx \prod_{\topic{j}\in\cateset} \prod_{\doc{i}\in\docuset} \prod_{\phrase{k}\in\topicphs{j}\cap\doc{i}} P(\phrase{k}|\doc{i},\topic{j}) P(\doc{i}|\topic{j}).
    \end{split}
    \raisetag{49pt}
\end{equation}
In the end, the total likelihood is factorized into the topic-conditional likelihoods of a document and a phrase, i.e., $P(\doc{i}|\topic{j})$ and $P(\phrase{k}|\doc{i},\topic{j})$, for all the positive triples $(\topic{j}, \doc{i}, \phrase{k})$ collected from $\taxo$ and $\docuset$.
That is, each triple satisfies the condition that its phrase $\phrase{k}$ belongs to the topic $c_j$ (i.e., $p_k\in\topicphs{j}$) and also appears in the document $d_i$.

To maximize Equation~\eqref{eq:likelihood}, we propose a unified model for estimating $P(\doc{i}|\topic{j})$ and $P(\phrase{k}|\doc{i},\topic{j})$ via the tasks of \textit{topic-document similarity prediction} and \textit{topic-conditional phrase generation}, respectively.
In Figure~\ref{fig:framework} Left, for each positive triple $(\topic{j}, \doc{i}, \phrase{k})$, the former task increases the similarity between the topic $\topic{j}$ and the document $\doc{i}$.
This similarity indicates how confidently the document $\doc{i}$ includes any sentences or mentions about the topic $\topic{j}$.
At the same time, the latter task maximizes the decoding probability of the phrase $\phrase{k}$ (i.e., generates the phrase) 
conditioned on the topic $\topic{j}$ and the document $\doc{i}$.
The model parameters are jointly optimized for the two tasks, and each of them will be discussed in Section~\ref{subsec:training}.

\smallsection{Expansion Step}
\label{subsubsec:expansion}
\proposed expands the topic taxonomy by discovering novel topics and inserting them into the taxonomy.
To this end, it utilizes the trained model to generate the phrases $\phrase{}$ that have a high topic-conditional likelihood $P(\phrase{}|\vtopic{};\docuset)$ for a new topic $\vtopic{}$ from a given corpus $\docuset$.
In Figure~\ref{fig:framework} Right, it first places a virtual topic node $\vtopic{j}$ at a \textit{valid} insertion position in the hierarchy (i.e., a child position of a topic node $\topic{j}$), and then it collects the phrases relevant to the virtual topic by generating them from documents $\doc{i}\in\docuset$.
Finally, it identifies multiple novel topics by clustering the collected phrases into semantically coherent but distinguishable clusters, which are inserted as the new topic nodes at the position of the virtual node.
The details will be presented in Section~\ref{subsec:expansion}.

\subsection{Encoder Architectures}
\label{subsec:training}
For modeling the two likelihoods $P(\doc{i}|\topic{j})$ and $P(\phrase{k}|\doc{i},\topic{j})$, we introduce a topic encoder and a document encoder, which respectively computes the representation of a topic $\topic{j}$ and a document $\doc{i}$.

\begin{figure}[t]
    \centering
    \includegraphics[width=\linewidth]{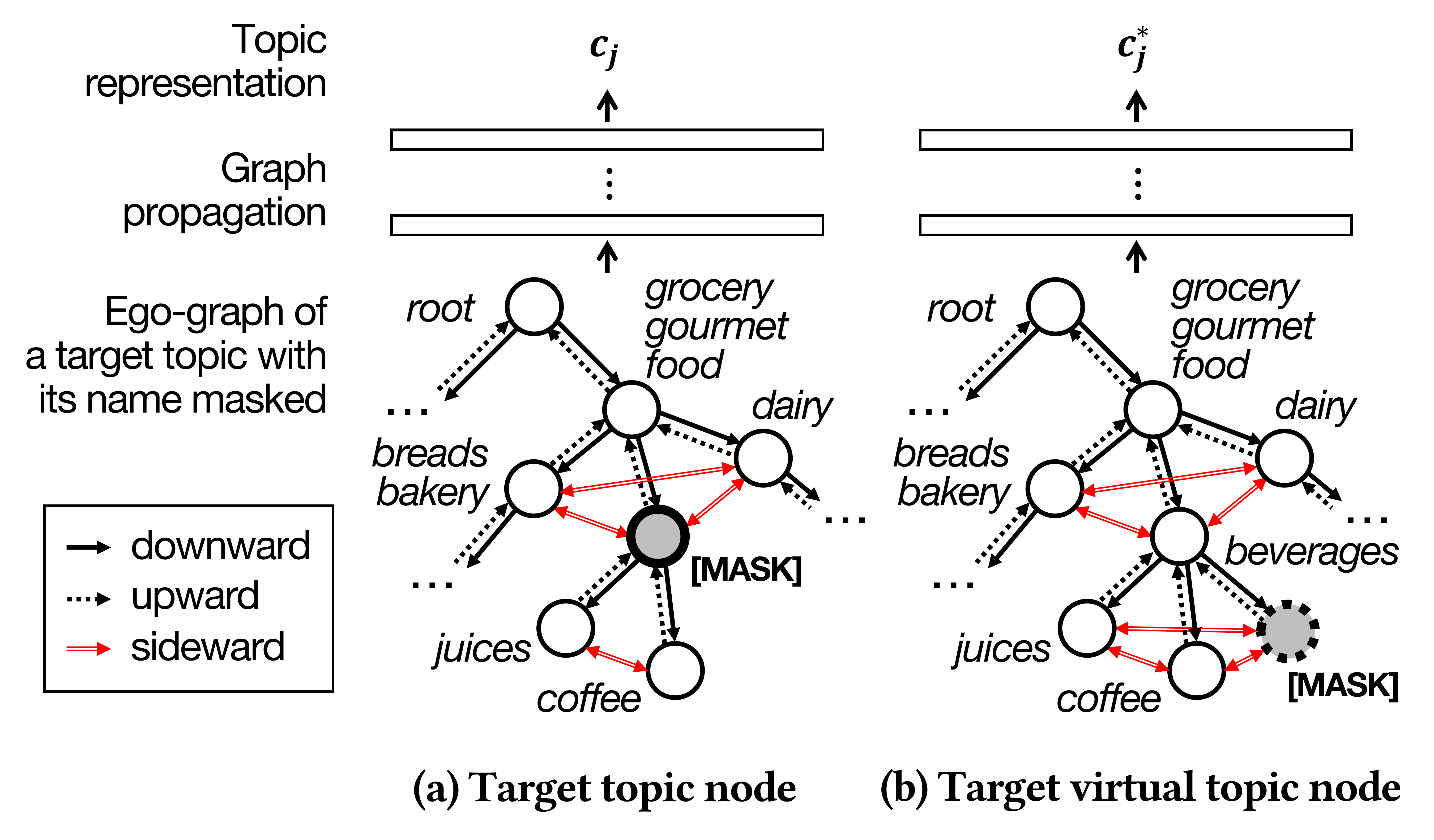}
    \caption{\reducetxt{The topic encoder architecture.
    It computes topic representations by encoding a topic relation graph.}
    }
    \label{fig:topic_encoder}
\end{figure}

\subsubsection{Topic Encoder}
\label{subsubsec:topic_encoder}
There are two important challenges of designing the architecture of a topic encoder:
(1) The topic encoder should be \textit{hierarchy-aware} so that the representation of each topic can accurately encode the hierarchical relation with its neighbor topics, and
(2) the representation of each topic needs to be \textit{discriminative} so that it can encode semantics distinguishable from that of the sibling topics.
Hence, we adopt graph convolutional networks (GCNs)~\cite{kipf2017semi} to capture the semantic relation structure surrounding each topic.

We first construct a topic relation graph $\mathcal{G}$ by enriching the edges of the given hierarchy $\taxo$ to model heterogeneous relations between topics, as shown in Figure~\ref{fig:topic_encoder}.
The graph contains three different types of inter-topic relations:
(1) downward, (2) upward, and (3) sideward.
The downward and upward edges respectively capture the top-down and bottom-up relations (i.e., hierarchy-awareness).
We additionally insert the sideward edges between sibling nodes that have the same parent node.
Unlike the downward and upward edges, the sideward edges pass the information in a negative way to make topic representations discriminative among the sibling topics.
The topic representation of $\topic{j}$ at the $m$-th GCN layer is computed by
\begin{equation}
\small
    \hvec{j}^{(m)}= \phi\left(
    \sideset{}{_{(i, j)\in\mathcal{G}}}\sum \alpha_{r(i, j)} \cdot \bm{W}^{(m-1)}_{r(i, j)} \cdot \hvec{i}^{(m-1)}\right),
\end{equation}
where $\phi$ is the activation function, $r(i, j)\in\{\text{down}, \text{up}, \text{side}\}$ represents the relation type of an edge $(i,j)$, and $\alpha$ indicates either positive or negative aggregation according to its relation type; 
i.e., $\alpha_{\text{down}}=\alpha_{\text{up}}=+1$ and $\alpha_{\text{side}}=-1$.
The \glove word vectors~\cite{pennington2014glove} for each topic name are used as the base node features (i.e., $\hvec{j}^{(0)}$) after being averaged for all tokens in the topic name.
Using a stack of $M$ GCN layers, we finally obtain the representation of a \textit{target} topic node $\topic{j}$ (i.e., the topic node that we want to obtain its representation) by $\cvec{j}=\hvec{j}^{(M)}$.

The topic encoder should also be able to obtain the representation of a virtual topic node, whose topic name is not determined yet, during the expansion step. 
For this reason, we mask the base node features of a target topic node regardless of whether the node is virtual or not, as depicted in Figures~\ref{fig:topic_encoder}(a) and (b).
In other words, with the name of a target topic masked, the topic representation encodes the relation structure of its $M$-hop neighbor topics.

\subsubsection{Document Encoder}
\label{subsubsec:document_encoder}
For the document encoder, we employ a pretrained language model, BERT~\cite{devlin2019bert}.
It models the interaction among the tokens based on the self-attention mechanism, thereby obtaining each token's contextualized representation, denoted by $[\vvec{i1},\ldots,\vvec{iL}]$.
A document representation $\dvec{i}$ is obtained by mean pooling in the end. 

\subsection{Learning Topic Taxonomy}
\label{subsec:training}
In the training step, \proposed optimizes model parameters by using positive triples as training data $\mathcal{X}=\{(\topic{j}, \doc{i}, \phrase{k})|\phrase{k}\in\topicphs{j}\cap\doc{i}, \forall \topic{j}\in\cateset, \forall\doc{i}\in\docuset\}$ 
via multi-task learning of {topic-document similarity prediction} and {topic-conditional phrase generation} (Sections~\ref{subsubsec:prediction} and \ref{subsubsec:generation}).

\subsubsection{Topic-Document Similarity Prediction}
\label{subsubsec:prediction}
The first task is to learn the similarity between a topic and a document.
We define the topic-document similarity score by bilinear interaction between their representations, i.e., $\cvec{j}^\top \bm{M} \dvec{i}$ where $\bm{M}$ is the trainable interaction matrix.
The topic-conditional likelihood of a document in Equation~\eqref{eq:likelihood} is optimized by using this topic-document similarity score, $P(\doc{i}|\topic{j}) \propto \exp(\cvec{j}^\top \bm{M} \dvec{i})$.

The loss function is defined based on InfoNCE \cite{oord2018representation}, which pulls positively-related documents into the topic while pushing away negatively-related documents from the topic.
\begin{equation}
\small
\label{eq:simloss}
    \simloss = - \sum_{(\topic{j},\doc{i},\phrase{k})\in\mathcal{X}} \log \frac{\exp(\cvec{j}^\top\bm{M} \dvec{i}/\gamma)}{\sum_{i'}\exp(\cvec{j}^\top \bm{M} \dvec{i'}/\gamma)},
\end{equation}
where $\gamma$ is the temperature parameter.
For each triple $(\topic{j}, \doc{i}, \phrase{k})$, we use its document $\doc{i}$ as positive and regard documents from all the other triples in the current mini-batch as negatives.

\subsubsection{Topic-Conditional Phrase Generation}
\label{subsubsec:generation}
The second task is to generate phrases from a document being conditioned on a topic.
For the phrase generator, we employ the architecture of the transformer decoder~\cite{vaswani2017attention}.

For topic-conditional phrase generation, the \textit{context} representation, $\bm{Q}(\topic{j}, \doc{i})$, needs to be modeled by fusing the textual content of a document $\doc{i}$ as well as the relation structure of a topic $\topic{j}$.
To leverage the textual features while focusing on the topic-relevant tokens, we compute \textit{topic-attentive} token representations and pass them as the input context of the transformer decoder.
Precisely, the topic-attention score of the $l$-th token in the document $\doc{i}$, $\beta_l(\topic{j},\doc{i})$, is defined by its similarity with the topic.
\vspace{-15pt}
\begin{equation}
\small
\label{eq:gencontext}
    \begin{split}
       \beta_l(\topic{j},\doc{i}) &= {\exp(\cvec{j}^\top\bm{M}\vvec{il})}/{\sideset{}{_{l'=1}^L}\sum\exp(\cvec{j}^\top\bm{M}\vvec{il'})} \\
        \bm{Q}(\topic{j},\doc{i}) &= [\beta_1(\topic{j},\doc{i}) \cdot \vvec{i1}, \ldots, \beta_{L}(\topic{j},\doc{i}) \cdot \vvec{iL}],
    \end{split}
\end{equation}
where the interaction matrix $\bm{M}$ is weight-shared with the one in Equation~\eqref{eq:simloss}.
Then, the sequential generation process of a token $\hat{v}_t$ is described by
\begin{equation}
\small
\label{eq:genoutput}
\begin{split}
    \svec{t} &= \text{Decoder}(\hat{v}_{<t};\bm{Q}(\topic{j},\doc{i})) \\ 
    \hat{v}_{t} &\sim \text{Softmax}(\text{FFN}(\svec{t})).
\end{split}
\end{equation}
$\text{FFN}$ is the feed-forward networks for mapping a state vector $\svec{t}$ into vocabulary logits.
Starting from the first token \texttt{[BOP]}, the phrase is acquired by sequentially decoding a next token $\hat{v}_t$ until the last token \texttt{[EOP]} is obtained;
the two special tokens indicate the begin and the end of the phrase.


The loss function is defined by the negative log-likelihood, where the phrase $\phrase{k}=[\token{k1},\ldots,\token{kT}]$ in a positive triple $(\topic{j},\doc{i},\phrase{k})$ is used as the target sequence of word tokens.
\begin{equation}
\small
\label{eq:genloss}
    \begin{split}
        \genloss &= - \sum_{(\topic{j}, \doc{i}, \phrase{k})\in\mathcal{X}} \sum_{t=1}^{T}  \ \log P(\token{kt}|\token{k(<t)},\topic{j},\doc{i}). \\
    \end{split}
\end{equation}

To sum up, the joint optimization of Equations~\eqref{eq:simloss} and \eqref{eq:genloss} updates all the model parameters in an end-to-end manner, including the similarity predictor, the phrase generator, and both encoders.

\subsection{Expanding Topic Taxonomy}
\label{subsec:expansion}
In the expansion step, \proposed expands the topic taxonomy by utilizing the trained model to generate the phrases 
for a virtual topic, which is assumed to be located at a valid insertion position in the hierarchy.
For thorough expansion, it considers a child position of every existing topic node as the valid position.
That is, for each virtual topic node $\vtopic{j}$ (referring to a new child of a topic node $\topic{j}$) {one at a time}, it performs topic phrase generation and clustering (Sections~\ref{subsubsec:collection} and~\ref{subsubsec:clustering}) to discover multiple novel topic nodes at the position.

\subsubsection{Novel Topic Phrase Generation}
\label{subsubsec:collection}
Given a virtual topic node $\vtopic{j}$ and each document $\doc{i}\in\docuset$, the trained model computes the topic-document similarity score and generates a topic-conditional phrase 
$\vphrase{}=[\hat{v}_{1},\ldots,\hat{v}_{T}]$ where $\hat{v}_{t}\sim P(\token{t}|\hat{v}_{<t},\vtopic{j},\doc{i})$.
Here, the generated phrase $\vphrase{}$ is less likely to belong to the virtual topic $\vtopic{j}$ if its source document $\doc{i}$ is less relevant to the virtual topic.
Thus, we utilize the topic-document similarity score as the \textit{confidence} of the generated phrase.
To collect only qualified topic phrases, we filter out non-confident phrases whose normalized topic-document similarity is smaller than a threshold, i.e., $P(\doc{i}|\vtopic{j})\approx\text{Norm}_{\doc{i}\in\docuset}(\exp(\cvec{j}^{*\top} \bm{M} \dvec{i})) < \tau$.
In addition to the confidence-based filtering, we exclude phrases that do not appear in the corpus at all, since they are likely implausible phrases.
This has substantially reduced the \textit{hallucination} problem of a generation model.

\subsubsection{Novel Topic Phrase Clustering}
\label{subsubsec:clustering}
To identify multiple novel topics at the position of the virtual topic node $\vtopic{j}$, we perform clustering on the phrases collected for the virtual topic.
We acquire semantic features of each phrase by averaging the \glove vectors~\cite{pennington2014glove} of word tokens in the phrase, then run $k$-means clustering with the initial number of clusters $k$ manually set.
Among the clusters, we selectively identify the new topics based on their cluster size, and the center phrase of each cluster is used as the topic name.

\section{Experiments}
\label{sec:exp}

\subsection{Experimental Settings}
\label{subsec:expsetting}
\smallsection{Datasets}
\label{subsubsec:dataset}
We use two real-world document corpora with their three-level topic taxonomy: \textbf{\amazon}~\cite{mcauley2013hidden} contains product reviews collected from Amazon, and \textbf{\dbpedia}~\cite{lehmann2015dbpedia} contains Wikipedia articles.
All the documents in both datasets are tokenized by the BERT tokenizer~\cite{devlin2019bert} and truncated to have maximum 512 tokens.
The statistics are listed in Table~\ref{tbl:datastats}.

\smallsection{Baseline Methods}
\label{subsubsec:baseline}
We consider methods for building a topic taxonomy from scratch, \textbf{\hlda} \cite{blei2003hierarchical} and \textbf{\taxogen} \cite{zhang2018taxogen}. 
We also evaluate the state-of-the-art methods for topic taxonomy expansion,
\textbf{\corel} \cite{huang2020corel} and \textbf{\taxocom} \cite{lee2022taxocom}.\footnote{The implementation details and hyperparameter selection for the baselines and \proposed are in Sections~\ref{subsec:basedetail} and~\ref{subsec:implementation}.}
Both of them identify and insert new topic nodes based on term embedding and clustering, with the initial topic taxonomy leveraged as supervision.

\smallsection{Experimental Settings}
To evaluate the performance for novel topic discovery, 
we follow the previous convention that randomly deletes half of leaf nodes from the original taxonomy and asks each expansion method to reproduce them~\cite{shen2020taxoexpan, lee2022taxocom}.
Considering the deleted topics as \textit{ground-truth}, we measure how completely new topics are discovered and how accurately they are inserted into the taxonomy.


\begin{table}[t]
\small
\caption{The statistics of the datasets.}
\label{tbl:datastats}
\centering
\resizebox{0.99\linewidth}{!}{%
\begin{tabular}{cccc}
    \toprule
        \textbf{Corpus} & \textbf{Vocab. size} & \textbf{\# Documents} & \textbf{\# Topic nodes} \\\midrule
        \amazon & 19,615 & \ \ 29,487 & 531 \\
        \dbpedia & 27,435 & 196,665 & 298 \\
    \bottomrule
\end{tabular}
}
\end{table}

\begin{table*}[t]
\caption{Quantitative evaluation on output topic taxonomies. 
The average and standard deviation for the three aspects are reported. 
The relation accuracy and subtopic integrity are considered only for the expansion methods, whose identified new topic nodes can be clearly compared with the ground-truth ones at each valid position.
}
\centering
\resizebox{0.99\linewidth}{!}{%
\begin{tabular}{ccPPPPPP}
    \toprule
     \multirow{2.5}{*}{\shortstack{\textbf{Part}}} & \multirow{2.5}{*}{\shortstack{\textbf{Methods}}} & \multicolumn{3}{c}{\textbf{\amazon}} & \multicolumn{3}{c}{\textbf{\dbpedia}} \\\cmidrule(lr){3-5}\cmidrule(lr){6-8}
    & & {{\small {\textbf{Term Coherence}}}} & {{\small {\textbf{Relation Accuracy}}}} & {{\small {\textbf{Subtopic Integrity}}}} & {{\small \textbf{Term Coherence}}} & {{\small \textbf{Relation Accuracy}}} & {{\small \textbf{Subtopic Integrity}}} \\\midrule
    
    \multirow{2}{*}{}
    & \hlda & 0.2417  (0.0398) &  N/A & N/A & 0.2688  (0.0320) &  N/A &  N/A \\
    & \taxogen & 0.4333  (0.1062) &  N/A &  N/A & 0.4906  (0.1523) &  N/A & N/A \\\midrule
    
    \multirow{3}{*}{$\subtaxo{1}$}
    & \corel & 0.5167  (0.1512) & 0.4833  (0.1501) & 0.2708  (0.1263) & 0.5083  (0.1377) & 0.6583  (0.1762) & 0.2813  (0.1727) \\
    & \taxocom & 0.6667  (0.1411) & 0.5167  (0.0992) & 0.3177  (0.1006) & 0.5250  (0.2151) & 0.6833  (0.1808) & 0.2917  (0.1282) \\
    \rowcolor{Gray} \cellcolor{white} 
    & \proposed & \textbf{0.9750}  (0.0496) & \textbf{0.8833}  (0.1113) & \textbf{0.4948}  (0.1309) & \textbf{0.9667}  (0.0713) & \textbf{0.9333}  (0.0713) & \textbf{0.5781}  (0.1389)  \\\midrule
    
    \multirow{3}{*}{$\subtaxo{2}$}
    & \corel & 0.5583  (0.1967) & 0.6333  (0.1594) & 0.2569  (0.1215) & 0.4417  (0.1815) & 0.5583  (0.1231) & 0.1458  (0.1488) \\
    & \taxocom & 0.6083  (0.1466) & 0.6167  (0.1369) & 0.4514  (0.1464) & 0.4833  (0.1944) & 0.7083  (0.1467) & 0.2708  (0.1282)\\
    \rowcolor{Gray} \cellcolor{white} 
    & \proposed & \textbf{0.8917}  (0.0707) & \textbf{0.8583}  (0.1650) & \textbf{0.6597}  (0.2062) & \textbf{0.9583}  (0.0707) & \textbf{0.9167}  (0.1054) & \textbf{0.5729}  (0.1035) \\\midrule
    
    \multirow{3}{*}{$\subtaxo{3}$} 
    & \corel & 0.5667  (0.1638) & 0.5833  (0.1222) & 0.2344  (0.1527) & 0.6250  (0.1669) & 0.7167  (0.1321) & 0.3177  (0.1195) \\
    & \taxocom & 0.5917  (0.1571) & 0.6083  (0.0972) & 0.1563  (0.1179) & 0.5667  (0.1852) & 0.6917  (0.1179) & 0.4167  (0.1069) \\
    \rowcolor{Gray} \cellcolor{white} 
    & \proposed & \textbf{0.9167}  (0.0690) & \textbf{0.9083}  (0.1004) & \textbf{0.4531}  (0.1068) & \textbf{0.9833}  (0.0309) & \textbf{0.9417}  (0.0904) & \textbf{0.6719}  (0.0916) \\\bottomrule
\end{tabular}
}
\label{tbl:humaneval}
\end{table*}

\begin{table}[t]
\caption{Performance for topic phrase generation.}
\centering
\resizebox{0.99\linewidth}{!}{%
\begin{tabular}{lrrrr}
    \toprule
    \multirow{2.5}{*}{\textbf{Methods}} & \multicolumn{2}{c}{\textbf{\amazon}} & \multicolumn{2}{c}{\textbf{\dbpedia}} \\\cmidrule(lr){2-3}\cmidrule(lr){4-5}
    & \textbf{PPL} $\downarrow$ & \textbf{ACC} $\uparrow$ & \textbf{PPL} $\downarrow$ & \textbf{ACC} $\uparrow$ \\\midrule
    \rowcolor{Gray}
    \proposed & \textbf{5.2553} & \textbf{0.6958}  & \textbf{3.1108} & \textbf{0.7768} \\
    {\small \ \ (Encoder) BERT$\rightarrow$Bi-GRU} & 5.7844 & 0.6884 & 3.5322 & 0.7645 \\
    {\small \ \ (Decoder) Transformer$\rightarrow$GRU} & 6.6649 & 0.6754 & 5.3690 & 0.6798 \\
    {\small \ \ w/o Topic-attentive context} & 7.0907 & 0.6643 & 7.1679 & 0.6553\\
    {\small \ \ w/o Hierarchical topic relation} & 6.5345 & 0.6772 & 3.9802 & 0.7423 \\
    {\small \ \ w/o Sideward topic relation} & 5.8705 & 0.6807 & 3.6985 & 0.7506 \\
    \midrule
    
    \multicolumn{2}{l}{{\footnotesize \textrank~\cite{mihalcea2004textrank}}\hfill -} & 0.3023 & \multicolumn{1}{c}{$~~~$-} & 0.1628 \\
    \multicolumn{2}{l}{{\footnotesize \topicrank~\cite{bougouin2013topicrank}}\hfill $~~~~$-} & 0.2099 & \multicolumn{1}{c}{$~~~$-} & 0.1092 \\
    {\small \topickg~\cite{wang2019topic}} & 13.1298 & 0.2770 & 11.5663 & 0.3238 \\
    {\small \bertkg~\cite{liu2020keyphrase}} & 11.0229 & 0.4165 & 9.4723 & 0.4734 \\
    {\small \berttkg~\cite{zhou2021topic}} & 10.9746 & 0.4308 & 8.3607 & 0.4894 \\
    \bottomrule
\end{tabular}
}
\label{tbl:genperf}
\end{table}

\subsection{Quantitative Evaluation}
\subsubsection{Topic Taxonomy Expansion}
\label{subsubsec:humaneval1}
First of all, we assess the quality of output topic taxonomies.
Following previous topic taxonomy evaluations~\cite{huang2020corel, lee2022taxocom}, we recruit 10 doctoral researchers and use their domain knowledge to examine three different aspects of a topic taxonomy.
\textbf{Term coherence} indicates how strongly terms in a topic node are relevant to each other. 
\textbf{Relation accuracy} computes how accurately a topic node is inserted into the topic taxonomy (i.e., \textit{precision} for novel topic discovery).
\textbf{Subtopic integrity} measures the completeness of subtopics for a topic node (i.e., \textit{recall} for novel topic discovery).
For exhaustive evaluation, we divide the output taxonomy of each expansion method into three disjoint parts $\subtaxo{1}$, $\subtaxo{2}$, and $\subtaxo{3}$ so that each of them covers some first-level topics (and their subtrees) in Table~\ref{tbl:taxopart} in Section~\ref{subsec:evalprotocol}.\footnote{The details of the evaluation protocol are in Section~\ref{subsec:evalprotocol}.}

In Table~\ref{tbl:humaneval}, \proposed achieves the highest scores for all the aspects.\footnote{The evaluation results obtain the Kendall coefficient of 0.96/0.91/0.84 (\amazon) and 0.93/0.90/0.91 (\dbpedia) respectively for each aspect, which indicates strong inter-rater agreement on ranks of the methods.}
For all the baseline methods, the term coherence is not good enough~because they assign candidate terms into a new topic according to the topic-term relevance mostly learned~from term co-occurrences.
In contrast, \proposed effectively collects coherent terms relevant to a new topic (i.e., term coherence $\geq 0.90$) by directly generating the topic-conditional terms from documents.
\proposed also shows significantly higher relation accuracy and subtopic integrity than the other expansion methods, with the help of its GNN-based topic encoder that captures a holistic topic structure beyond the first-order topic relation.

\begin{table*}[t]
\caption{Top-5 topic terms included in each topic node. The off-topic (or too general) terms are marked with a strikethrough, and the multi-word terms that are not obtainable by the extraction-based methods are underlined.}
\centering
\setlength{\tabcolsep}{4pt}
\resizebox{\linewidth}{!}{%
\begin{tabular}{cQR}
    \toprule
    & \multicolumn{1}{l}{\textbf{\amazon}} & \multicolumn{1}{l}{\textbf{\dbpedia}} \\\midrule
    
    \textbf{Topic}
    & \reduce{\textbf{Root $\rightarrow$ grocery gourmet food $\rightarrow$ beverages $\rightarrow$ tea}}
    & \reduce{\textbf{Root $\rightarrow$ work $\rightarrow$ periodical literature $\rightarrow$ magazine}} \\\cmidrule(lr){1-1}\cmidrule(lr){2-2}\cmidrule(lr){3-3}
    
    \reduce{\corel}
    & \reduce{blend, alvita, green tea, white tea, herbal tea}
    & \reduce{science fiction, comics, anthology, \wrong{monthly}, newspaper} \\
    
    \reduce{\taxocom}
    & \reduce{earl grey, taylors of harrogate, black tea, herbal tea, teabags} 
    & \reduce{\wrong{books}, cover, fashion, \wrong{specializing}, publishers} \\
    
    \reduce{\proposed}
    & \reduce{green tea, herbal tea, black tea, \unique{vanilla chai tea}, \unique{jasmine tea}}
    & \reduce{news mag.., monthly mag.., \unique{business mag..}, \unique{comics mag..}, \unique{fashion mag..}} \\\midrule
    
    \textbf{Topic}
    & \reduce{\textbf{Root $\rightarrow$ personal care $\rightarrow$ household supplies $\rightarrow$ dishwashing}}
    & \reduce{\textbf{Root $\rightarrow$ event $\rightarrow$ societal event $\rightarrow$ film festival}} \\\cmidrule(lr){1-1}\cmidrule(lr){2-2}\cmidrule(lr){3-3}
    
    \reduce{\corel} 
    & \reduce{\wrong{household}, fresh, flushes, dirty, \wrong{biological}} 
    & \reduce{film festival, \wrong{music festival}, annual event, \wrong{festival}, cinema} \\
    
    \reduce{\taxocom}
    & \reduce{dishwasher, bubbles, dirty, smelly, nasty}
    & \reduce{film festival, \wrong{fest}, \wrong{festival}, independent film, annual festival} \\
    
    \reduce{\proposed}
    & \reduce{dishwasher, dish soap, \unique{dish detergent}, \unique{dishwasher gel}, \unique{dishwasher soap}} 
    & \reduce{film festi.., \unique{short film festi..}, annual film festi.., \unique{cannes film festi..}, \unique{venice film festi..}}\\\midrule
    
    \textbf{Topic}
    & \reduce{\textbf{Root $\rightarrow$ baby products $\rightarrow$ diapering $\rightarrow$ cloth diapers}}
    & \reduce{\textbf{Root $\rightarrow$ sports season $\rightarrow$ sports team season $\rightarrow$ ncaa team season}} \\\cmidrule(lr){1-1}\cmidrule(lr){2-2}\cmidrule(lr){3-3}
    
    \reduce{\corel}
    & \reduce{cloth diaper, \wrong{diapers}, \wrong{diaperbag}, cloth diapering, {baby wipes}}
    & \reduce{naia, div.. ii, ncaa div.. ii, atlantic coast conference, mid american conference} \\
    
    \reduce{\taxocom}
    & \reduce{\wrong{disposable diapers}, \wrong{biodegradeable}, cloth diaper, \wrong{diapering}, bulky}
    & \reduce{mastodons, basketball tournament, college football, bulldogs, hoosiers} \\
    
    \reduce{\proposed}
    & \reduce{cloth diaper, prefold diaper, \unique{diaper cover}, \unique{diaper liner}, \unique{reusable diaper}}
    & \reduce{\unique{ncaa national team}, ncaa tournament, ncaa div.. ii, ncaa div.. i, \unique{ncaa championship}} \\

    \bottomrule
\end{tabular}
}
\label{tbl:topicterms}
\end{table*}

\begin{table*}[t]
\caption{Novel topics identified at each target position. The center term (i.e., topic name) of each identified topic is presented. Correct topics ($\Circle$), incorrect topics ($\otimes$), and redundant topics (\ding{34}) are annotated.}
\setlength{\tabcolsep}{4pt}
\centering
\resizebox{\linewidth}{!}{%
\begin{tabular}{cSW}
    \toprule
    & {\textbf{\amazon}} &  {\textbf{\dbpedia}} \\ \midrule
    
    \reduce{\textbf{Position}}
    & \reduce{\textbf{Root $\rightarrow$ grocery gourmet food $\rightarrow$ beverages $\rightarrow$ ?}} 
    & \reduce{\textbf{Root $\rightarrow$ agent $\rightarrow$ sports team $\rightarrow$ ?}}
    \\\cmidrule(lr){1-1}\cmidrule(lr){2-2}\cmidrule(lr){3-3}
    
    \reduce{\textbf{Sibling topics}}
    & \reduce{\textbf{tea, coffee, hot cocoa, water, sports drinks}} 
    & \reduce{\textbf{basketball team, cycling team, football team}}
    \\\cmidrule(lr){1-1}\cmidrule(lr){2-2}\cmidrule(lr){3-3}
    
    \reduce{\corel} 
    & \reduce{apple cider ($\Circle$), bottles ($\otimes$), drinking ($\otimes$), fruit juice ($\Circle$), matcha (\ding{34})} 
    & \reduce{baseball ($\otimes$), domestic competition ($\otimes$), football club (\ding{34}), national~ice~hockey~team ($\Circle$), soccer ($\otimes$)} \\
    
    \reduce{\taxocom} 
    & \reduce{decaf chai (\ding{34}), espresso (\ding{34}), fizzy ($\Circle$), juice ($\Circle$), lipton~(\ding{34}), mouthwash ($\otimes$)} 
    & \reduce{hockey ($\otimes$), junior football team (\ding{34}), national team ($\Circle$), regular~season~($\otimes$), rugby~club ($\Circle$)} \\
    
    \reduce{\proposedab}
    & \reduce{breakfast tea (\ding{34}), coconut water ($\Circle$), fruit juice ($\Circle$), natural~cocoa (\ding{34}), vanilla coffee (\ding{34})} 
    & \reduce{american~football~team (\ding{34}), cricket~team ($\Circle$), cycling~team (\ding{34}), professional~basketball~team (\ding{34}), rugby~union~team ($\Circle$)} \\
    
    \reduce{\proposedfull}
    & \reduce{coconut water ($\Circle$), cream soda ($\Circle$), decaf tea (\ding{34}), diet~smoothie ($\Circle$), redline energy drink ($\Circle$)} 
    & \reduce{beach~handball~team ($\Circle$), cricket~team ($\Circle$), football club (\ding{34}), ice~hockey~team ($\Circle$), rugby~union~team ($\Circle$)} \\
    \bottomrule
\end{tabular}
}
\label{tbl:noveltopics}
\end{table*}

\subsubsection{Topic-Conditional Phrase Generation}
\label{subsubsec:ablation}
We investigate the topic phrase prediction performance of our framework and other keyphrase extraction/generation models.
We leave out 10\% of the positive triples $(\topic{j}, \doc{i}, \phrase{k})$ from the training set $\mathcal{X}$ and use them as the test set.
We measure \textbf{perplexity (PPL)} and \textbf{accuracy (ACC)} by comparing each generated phrase with the target phrase at the token-level and phrase-level, respectively.

In Table~\ref{tbl:genperf}, \proposed achieves the best PPL and ACC scores.
We observe that \proposed more accurately generates topic-related phrases from input documents, compared to the state-of-the-art keyphrase generation methods which are not able to consider a specific topic as the condition for generation.
In addition, ablation analyses validate that each component of our framework contributes to accurate generation of topic phrases. 
Particularly, the hierarchical (i.e., upward and downward) and sideward relation modeling of the topic encoder improves the quality of generated phrases.

\subsection{Qualitative Evaluation}
\subsubsection{Comparison of Topic Terms}
\label{subsubsec:topicterms}
We qualitatively compare the topic terms found by each method.
In case of \proposed, we sort all confident topic terms by their cosine distances to the topic name (i.e., center term) using the global embedding features~\cite{pennington2014glove}.

Table~\ref{tbl:topicterms} shows that the topic terms of \proposed are superior to those of the baseline methods, in terms of the expressiveness as well as the topic relevance.
In detail, some of the terms retrieved by \corel and \taxocom are either off-topic or too general (marked with a strikethrough);
this indicates that their topic relevance score for each term is not good at capturing the hierarchical topic knowledge of a text corpus.
On the contrary, \proposed generates strongly topic-related terms by capturing the relation structure of each topic.
Furthermore, \proposed is effective to find infrequently-appearing multi-word terms (underlined), which all the extraction-based methods fail to obtain.

\begin{figure}[t]
    \centering
    \includegraphics[width=\linewidth]{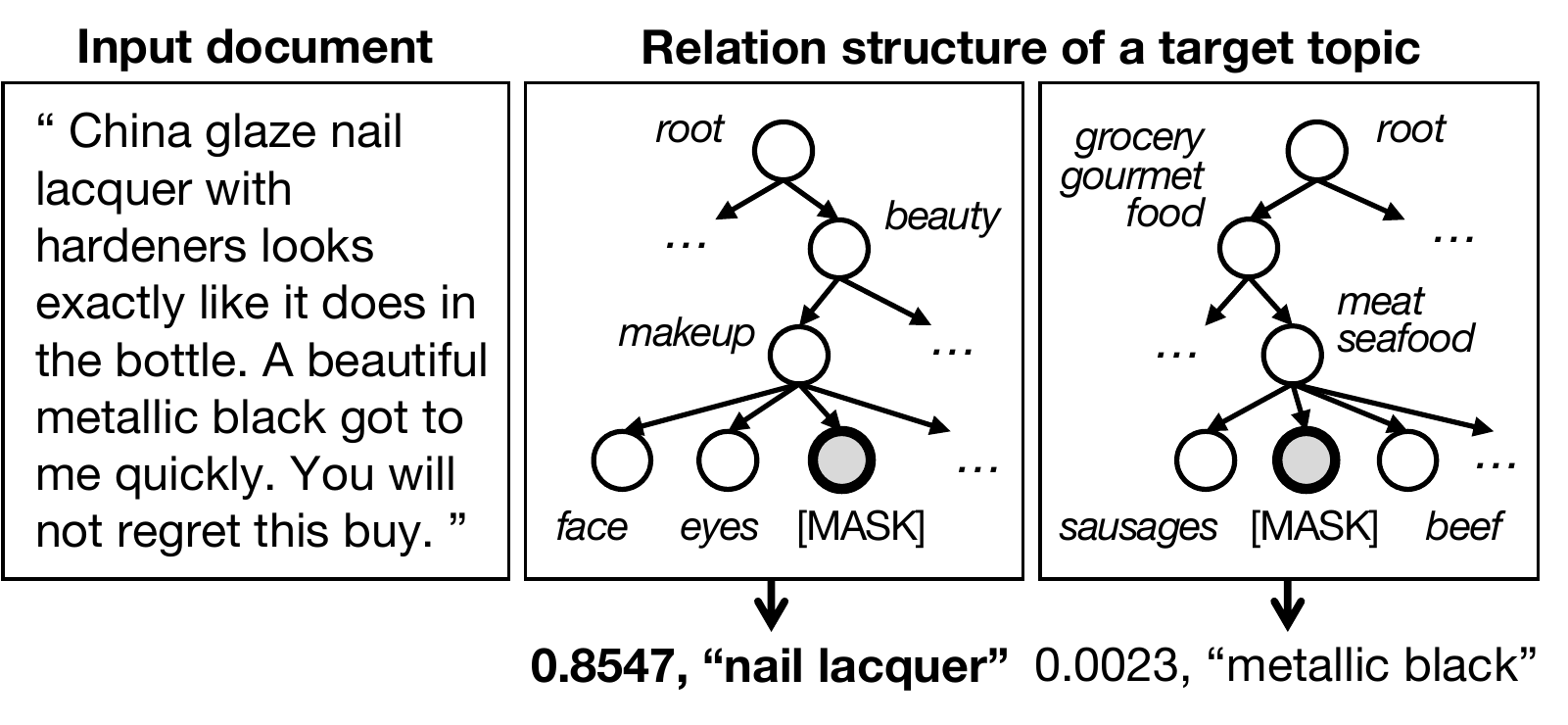}
    \caption{Examples of topic-conditional phrase generation, given a document and its relevant/irrelevant topic.}
    \label{fig:casestudy}
\end{figure}

\subsubsection{Comparison of Novel Topics}
\label{subsubsec:noveltopics}
Next, we examine novel topics inserted by each expansion method.
To show the effectiveness of sideward relation modeling adopted by our topic encoder (Section~\ref{subsubsec:topic_encoder}),
we additionally present the results of \textbf{\proposedfull} and \textbf{\proposedab}, which computes topic representations with and without capturing the \underline{s}ideward topic \underline{r}elations.

In Table~\ref{tbl:noveltopics}, \proposedfull successfully discovers new topics that should be placed in a target position.
Notably, the new topics are clearly distinguishable from the sibling topics (i.e., known topics given in the initial topic hierarchy), which reduces the redundancy of the output topic taxonomy.
On the other hand, \corel and \taxocom show limited performance for novel topic discovery; 
some new topics are redundant (\ding{34}) while some others do not preserve the hierarchical relation with the existing topics ($\otimes$).
Some of the new topics found by \proposedab semantically overlap with the sibling topics, even though they are at the correct position in the hierarchy;
this implies that our topic encoder with sideward relation modeling makes the representation of a virtual topic node \textit{discriminative} with its sibling topic nodes, and it eventually helps to discover new conceptual topics of novel semantics.

\subsubsection{Case Study of Topic Phrase Generation}
\label{subsubsec:casestudy}
To study how the generated phrases and their topic-document similarity scores (i.e., confidences) vary depending on a topic condition, we provide examples of topic-conditional phrase generation.
The input document in Figure~\ref{fig:casestudy} contains a review about nail care products. 
In case that the relation structure of a target topic implies the nail product (Figure~\ref{fig:casestudy} Left),
\proposed obtains the desired \textit{topic-relevant} phrase ``nail lacquer'' along with the high topic-document similarity of 0.8547.
On the other hand, given the relation structure of a target topic which is inferred as a kind of meat foods (Figure~\ref{fig:casestudy} Right),
it generates a \textit{topic-irrelevant} phrase ``metallic black'' from the document along with the low topic-document similarity of 0.0023.
That is, \proposed fails to get a qualified topic phrase when the textual contents of an input document is obviously irrelevant to a target topic. 
In this sense, \proposed filters out non-confident phrases having a low topic-document similarity score to collect only the phrases relevant to each virtual topic.

\subsection{Analysis of Topic-Document Similarity}
\label{subsec:unseenphs}
\begin{figure}[t]
    \centering
    \includegraphics[width=\linewidth]{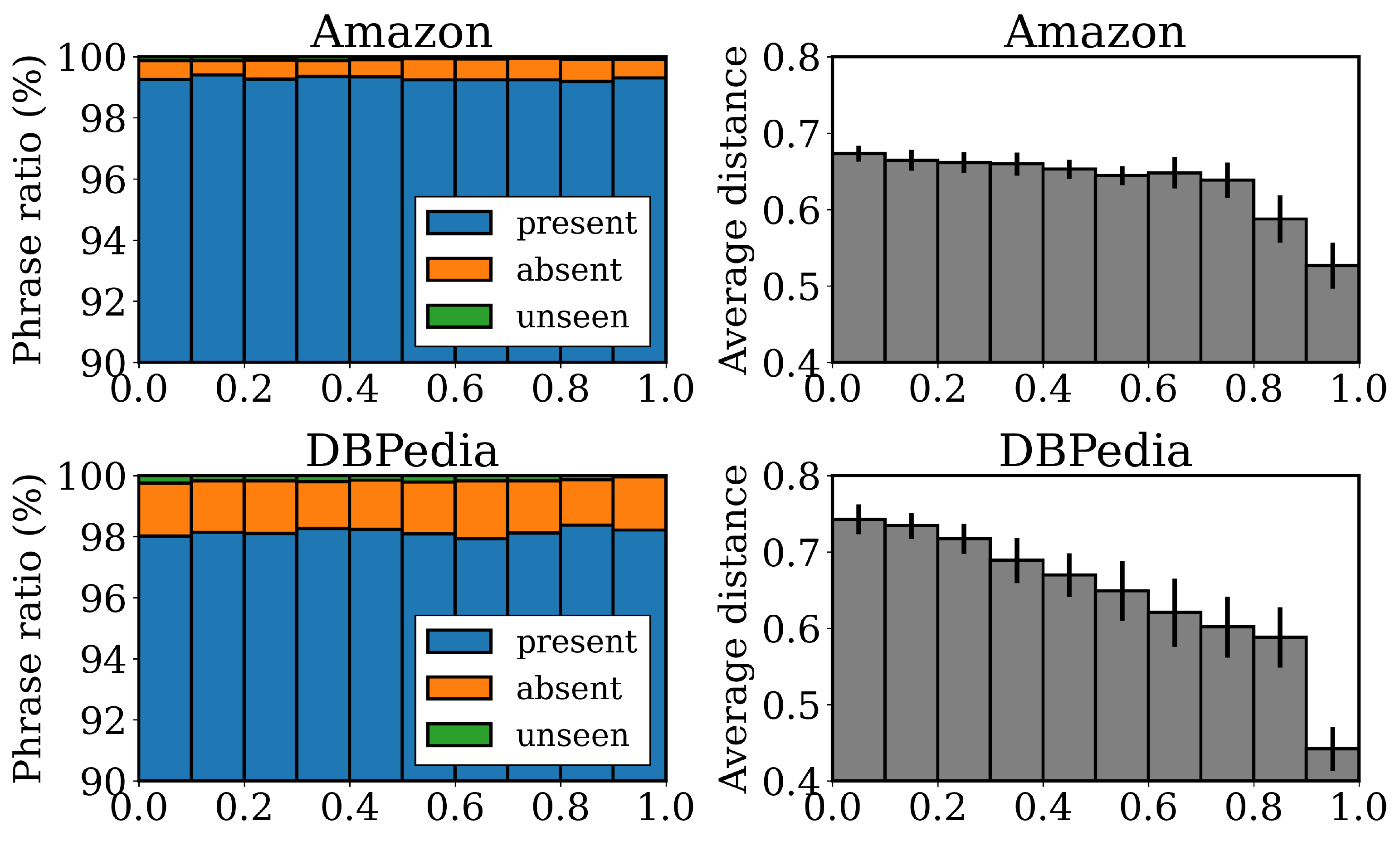}
    \caption{\reducetxt{The ratio of three categories for generated phrases (Left) and the average semantic distance among generated phrases (Right). The horizontal axis shows 10 bins of normalized topic-document similarity scores.}}
    \label{fig:confanal}
\end{figure}

Finally, we investigate the changes of generated phrases in two aspects, with respect to the topic-document similarity scores.
The first aspect is the ratio of three categories for generated phrases, which have been focused on in the literature of keyphrase generation~\cite{meng2017deep, zhou2021topic}: 
\textbf{(1) present phrases} appearing in the input document, 
\textbf{(2) absent phrases} not appearing in the input document but in the corpus at least once, 
and \textbf{(3) unseen (i.e., totally-new) phrases} that are not observed in the corpus at all.
The second aspect is the average semantic distance among the phrases, measured by using the semantic features. 
For the plots in Figure~\ref{fig:confanal}, the horizontal axis represents 10 bins of normalized topic-document similarity scores over all generated phrases.

Interestingly, \proposed hardly generates absent phrases (about $0.7\%$ for \amazon, $1.7\%$ for \dbpedia) and unseen phrases (about $0.1\%$ for \amazon, $0.2\%$ for \dbpedia) regardless of the topic-document similarity; instead, it generates present phrases in most cases (Figure~\ref{fig:confanal} Left). 
In other words, if the input document is not relevant to a target topic, it tends to generate an irrelevant-but-present phrase rather than a relevant-but-absent phrase, as shown in Section~\ref{subsubsec:casestudy}. 
One potential risk of \proposed is to generate unseen phrases that are nonsense or implausible, also known as \textit{hallucinations} in neural text generation, and such unseen phrases can degrade the quality and credibility of output topic taxonomies. 
This result supports that we can easily exclude all unseen phrases, which account for less than 0.2\% of generated phrases, to effectively address this issue.

Moreover, the negative correlation between the topic-document similarity score and the inter-phrase semantic distance (Figure~\ref{fig:confanal} Right) provides empirical evidence that the similarity score can serve as the confidence of a generated topic phrase.
There is a clear tendency toward decreasing the average semantic distance as the topic-document similarity score increases;
this implies that the phrases generated from topic-relevant documents are semantically coherent to each other, and accordingly, they are likely to belong to the same topic.

\section{Conclusion}
\label{sec:conc}
In this paper, we study the problem of topic taxonomy expansion, pointing out that the existing approach has shown limited term coverage and inconsistent topic relation.
Our \proposed framework introduces hierarchy-aware topic term generation, which generates a topic-related term by using both the textual content of an input document and the relation structure of a topic as the condition for generation.
The quantitative and qualitative evaluation demonstrates that our framework successfully obtains much higher-quality topic taxonomy in various aspects, compared to other baseline methods.  

For future work, it would be promising to incorporate an effective measure for the topic relevance of multi-word terms (i.e., phrases) into our framework.
Since learning and utilizing the representation of multi-word terms remains challenging and worth exploring, it can be widely applied to many other text mining tasks.

\section{Limitations}
\label{sec:limit}
Despite the remarkable performance of \proposed on our tested corpus, there is still room to improve regarding how to better handle topics, documents, and phrases, for effective mining of topic knowledge.
First, \proposed uses only the topic names (i.e., center terms) as the base node features in the topic relation graph, which makes our topic encoder difficult to capture the collective meaning of each topic from its set of topic-related phrases.
Second, the confidence of each generated phrase considers only the topic relevance of its source document, instead of all the documents in which this phrase appears.
Finally, the clustering process does not leverage the contextualized textual features computed by our BERT-based document encoder, which makes it hard to consolidate the context of the phrase within its source document.

\section*{Acknowledgements}
This work was supported by the IITP grant (No. 2018-0-00584, 2019-0-01906) and the NRF grant (No. 2020R1A2B5B03097210). 
It was also supported by US DARPA KAIROS Program (No. FA8750-19-2-1004), SocialSim Program (No. W911NF-17-C-0099), INCAS Program (No. HR001121C0165), National Science Foundation (IIS-19-56151, IIS17-41317, IIS 17-04532), and the Molecule Maker Lab Institute: An AI Research Institutes program (No. 2019897).

\bibliography{main}
\bibliographystyle{acl_natbib}

\newpage
\appendix
\section{Supplementary Material}
\subsection{Pseudo-code of \proposed}
\label{subsec:pseudocode}
Algorithm~\ref{alg:overview} describes the detailed process of our framework, including the training step (Lines 1--9) and the expansion step (Lines 10--23). 
The final output is the expanded topic taxonomy (Line 24).

\begin{algorithm}
\small
    \DontPrintSemicolon
    \SetKwProg{Fn}{Function}{:}{}
    \SetKwComment{Comment}{$\triangleright$\ }{} 
	
	\KwIn{Initial topic taxonomy $\taxo=(\cateset, \edgeset)$ and Text~corpus $\docuset$} 
	\KwOut{Expanded taxonomy $\taxo'$}
	
	\vspace{5pt}
	{\color{blue}{\tcp{Step 1: Learning the topic taxonomy}}}
	$\mathcal{X}\leftarrow$ \textsc{CollectTriples}$(\taxo, \docuset)$ \;
	$\mathcal{G} \leftarrow$ \textsc{ConstructGraph}$(\taxo)$\; 
    \While{not converged}{
    \For{$(\topic{j}, \doc{i}, \phrase{k}) \in \mathcal{X}$}{
        Obtain the model outputs for the inputs $(\mathcal{G}, \topic{j}, \doc{i})$ \;
        Compute $\mathcal{L}_{sim}$ by Equation~\eqref{eq:simloss} \;
        Compute $\mathcal{L}_{gen}$ by Equation~\eqref{eq:genloss} \;
        $\mathcal{L} \leftarrow \mathcal{L}_{sim}+\mathcal{L}_{gen}$ \;
        $\Theta \leftarrow \Theta - \eta\cdot {\partial\mathcal{L}}/{\partial\Theta}$ \;
        }
    }

    \vspace{5pt}
    {\color{blue}{\tcp{Step 2: Expanding the topic taxonomy}}}
    $\taxo' \leftarrow \taxo$ \;
    {\color{blue}{\tcp{For each valid position (the~child~position of each topic)}}}
    \For{$\topic{j}\in\cateset$}{
    $\taxo^*, \topicphs{}^* \leftarrow \taxo, \emptyset$ \;
    $\topic{j}^*\leftarrow$ \textsc{MakeVirtualNode}$(\topic{j})$ \;
    $\taxo^*$\textsc{.InsertNode}$(\topic{j}, \{\topic{j}^*\})$ \;
    $\mathcal{G}^* \leftarrow$ \textsc{ConstructGraph}$(\taxo^*)$ \; 
        \For{$\doc{i}\in\docuset$}{
        Obtain the model outputs for the inputs $(\mathcal{G}^*, \topic{j}^*, \doc{i})$ \;
        $\hat{s} \leftarrow \exp(\cvec{j}^{*\top}\bm{M}\dvec{i})$\;
        $\hat{p} \leftarrow [\hat{v}_{1},\ldots,\hat{v}_{T}], \hat{v}_{t}\sim P(\token{t}|\hat{v}_{<t},\doc{i},\vtopic{j})$\;
        
        $\topicphs{}^*$.\textsc{Append}$((\hat{s}, \hat{p}))$\;  
        }
        $\topicphs{}^*\leftarrow$ \textsc{FilterByNormalizedScore}$(\topicphs{}^*, \tau)$ \;
        $\topic{j1}^{*}, \ldots, \topic{jK}^{*} \leftarrow$ \textsc{ClusterPhrases}$(\topicphs{}^*)$ \;
        $\taxo'$.\textsc{InsertNode}$(\topic{j}, \{\topic{j1}^{*}, \ldots, \topic{jK}^{*}\})$ \;
    }

    \Return $\taxo'$
    
\caption{The process of \proposed.}
\label{alg:overview}
\end{algorithm}

\smallsection{Training Step (Lines 1--9)}
\proposed first collects all positive triples $(\topic{j}, \doc{i}, \phrase{k})$ from an initial topic taxonomy $\taxo$ and a text corpus $\docuset$ (Line 1; Section~\ref{subsubsec:training}), and constructs a topic relation graph $\mathcal{G}$ from the topic hierarchy (Line 2; Section~\ref{subsubsec:topic_encoder}).
Then, it updates all the trainable parameters based on the gradient back-propagation (Lines 5--9) to minimize the losses for the topic-document similarity prediction task (Line 6; Section~\ref{subsubsec:prediction}) and the topic-conditional phrase generation task (Line 7; Section~\ref{subsubsec:generation}).

\smallsection{Expansion Step (Lines 10--23)}
Using the trained model, \proposed discovers new topics that need to be inserted into each valid position in the topic hierarchy (Line 11).
For a virtual topic node $\topic{j}^*$ as a newly-introduced child of each topic node $\topic{j}$ (Line 13), it constructs a topic relation graph $\mathcal{G}^*$ from the topic hierarchy augmented with the virtual topic node (Lines 14--15).
Then, it collects all pairs of a topic-document similarity score and a generated topic phrase  $(\hat{s}, \hat{p})$, which are obtained by using the trained model on the augmented topic relation graph and all the documents (Lines 16--20; Section~\ref{subsubsec:collection}).
Next, it filters out non-confident (i.e., irrelevant) phrases according to the normalized score (Line 21), then it performs clustering to find out multiple phrase clusters, each of which is considered as a new topic node having a novel topic semantics (Line 22; Section~\ref{subsubsec:clustering}).
In the end, it inserts the identified new topic nodes into the target position (i.e., the child of a topic node $\topic{j}$) to expand the current topic taxonomy (Line 23).


\begin{figure}[t]
    \centering
    \includegraphics[width=\linewidth]{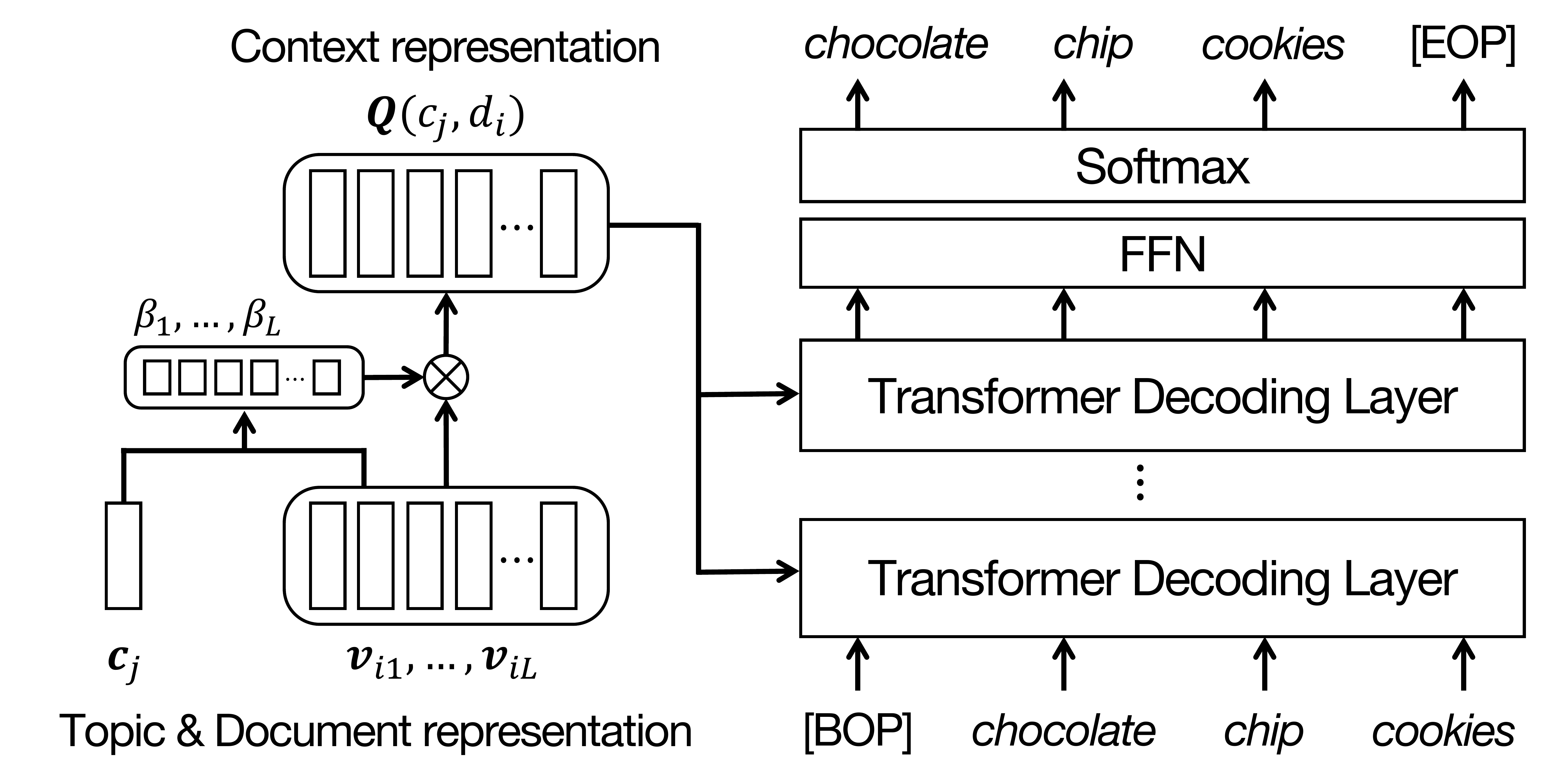}
    \caption{The phrase generator architecture. It generates the token sequence given a topic and a document, by using topic-attentive token representations as the context.
    }
    \label{fig:phrase_generator}
\end{figure}

\subsection{Baseline Methods}
\label{subsec:basedetail}
For the baselines, we employ the official author codes while following the parameter settings provided by~\cite{lee2022taxocom}.
For all the methods that optimize the Euclidean or spherical embedding space (i.e., \taxogen, \corel, and \taxocom), we fix the number of negative terms (for each positive term pair) to 2 during the optimization.
\begin{itemize}
    \item \textbf{\hlda}\footnote{https://github.com/joewandy/hlda}~\cite{blei2003hierarchical} performs hierarchical latent Dirichlet allocation.
    It models a document generation process as sampling its words along the path selected from the root to a leaf. 
    We set the smoothing parameters $\alpha$ = 0.1 and $\eta$ = 1.0, respectively for document-topic distributions and topic-word distributions, and the concentration parameter in the Chinese restaurant process $\gamma$ = 1.0. 
    
    \item \textbf{\taxogen}\footnote{https://github.com/franticnerd/taxogen}~\cite{zhang2018taxogen} is the unsupervised framework for topic taxonomy construction. 
    To identify hierarchical term clusters, it optimizes the term embedding space with SkipGram~\cite{mikolov2013distributed}.
    We set the maximum taxonomy depth to 3 and the number of child nodes to 5, as done in~\cite{zhang2018taxogen,shang2020nettaxo}.
    
    \item \textbf{\corel}\footnote{https://github.com/teapot123/CoRel}~\cite{huang2020corel} is the first topic taxonomy expansion method.
    It trains a topic relation classifier by using the initial taxonomy, then recursively transfers the relation to find out candidate terms for novel subtopics. 
    Finally, it identifies novel topic nodes based on term embeddings induced by SkipGram~\cite{mikolov2013distributed}.
    
    \item \textbf{\taxocom}\footnote{https://github.com/donalee/taxocom}~\cite{lee2022taxocom} is the state-of-the-art method for topic taxonomy expansion. 
    For each node from the root to the leaf, it recursively optimizes term embedding and performs term clustering to identify both known and novel subtopics.
    we set $\beta=1.5, 2.5, 3.0$ (for each level) in the novelty threshold $\tau_{nov}$, and fix the signficance threshold $\tau_{sig}=0.3$.
    
\end{itemize}

\subsection{Implementation Details}
\label{subsec:implementation}
\smallsection{Model Architecture}
For the topic encoder, we use two GCN layers to avoid the over-smoothing problem, and fix the dimensionality of all node representations to 300.
For the document encoder, we employ the \texttt{bert-base-uncased} provided by huggingface~\cite{devlin2019bert}, as the initial checkpoint of a pretrained model.
It contains 12 layers of transformer blocks with 12 attention heads, thereby obtaining 768-dimensional contextualized token representations $[\vvec{i1}, \ldots, \vvec{iL}]$ (and a final document representation $\dvec{i}=\text{mean-pooling}(\vvec{i1}, \ldots, \vvec{iL})$) for an input document $\doc{i}$.
Consequently, the size of the interaction matrix $\bm{M}$ in our topic-document similarity predictor (Equation~\eqref{eq:simloss}) becomes 300 $\times$ 768.
For the phrase generator, we adopt a single layer of the transformer decoder with 16 attention heads\footnote{We empirically found that the number of decoding layers hardly affects the performance (i.e., accuracy) of topic-conditional phrase generation.} and train its parameters from scratch without using the checkpoint of a pretrained text decoder.
We limit the maximum length of a generated phrase to 10.
Figure~\ref{fig:phrase_generator} shows the phrase generator architecture.
In total, our neural model contains 540K (for the topic encoder), 110M (for the document encoder), 230K (for the similarity predictor), and  30M (for the phrase generator) parameters.


\smallsection{Training Step}
For the optimization of model parameters, we use the Adam optimizer~\cite{kingma2014adam} with the initial learning rate 5e-5 and the weight decay 5e-6.
The batch size is set to 64, and the temperature parameter $\gamma$ in Equation~\eqref{eq:simloss} is set to 0.1.
The best model is chosen using the best perplexity of generated topic phrases on the validation set of positive triples $(\topic{j}, \doc{i}, \phrase{k})$, which is evaluated every epoch.

\smallsection{Expansion Step}
To filter out non-confident phrases (Section~\ref{subsubsec:collection}), we set the threshold value $\tau$ to 0.8 after applying min-max normalization on all topic-document similarity scores computed for each virtual topic node.
To perform $k$-means clustering on the collected topic phrases (Section~\ref{subsubsec:clustering}), we set the initial number of clusters $k$ to 10, then select top-5 clusters by their cluster size (i.e., the number of phrases assigned to each cluster).
The center phrase of each cluster is used as the final topic name of the new topic node.

\subsection{Computing Platform}
All the experiments are carried out on a Linux server machine with Intel Xeon Gold 6130 CPU @2.10GHz and 128GB RAM by using a single RTX3090 GPU.
In this environment, the model training of \proposed takes around 2 hours and 6 hours for \amazon and \dbpedia, respectively.

\begin{table}[b]
\small
\caption{Three disjoint parts of the topic taxonomy.}
\label{tbl:taxopart}
\centering
\resizebox{0.99\linewidth}{!}{%
\begin{tabular}{VcT}
    \toprule
        \textbf{Corpus} & \textbf{Part} & \textbf{First-level topics} \\\midrule
        & $\subtaxo{1}$ & grocery gourmet food, toys games\\
        \amazon & $\subtaxo{2}$ & beauty, personal care \\
        & $\subtaxo{3}$ & baby products, pet supplies \\\midrule
        & $\subtaxo{1}$ & agent, work, place \\
        \dbpedia & $\subtaxo{2}$ & species, unit of work, event \\
        & $\subtaxo{3}$ & sports season, device, topical concept \\
    \bottomrule
\end{tabular}
}
\end{table}

\subsection{Quantitative Evaluation Protocol}
\label{subsec:evalprotocol}
For exhaustive evaluation on a large-scale topic taxonomy with hundreds of topic nodes, the output taxonomy of topic taxonomy expansion methods (i.e., \corel, \taxocom, and \proposed) is divided into three parts $\subtaxo{1}$, $\subtaxo{2}$, and $\subtaxo{3}$ so that each part covers some of the first-level topics (and their subtrees) listed in Table~\ref{tbl:taxopart}.

In case of \hlda and \taxogen, the first-level topics in their output taxonomies are not matched with the ground-truth topics (in Table~\ref{tbl:taxopart}), because they build a topic taxonomy from scratch.
For this reason, in Table~\ref{tbl:humaneval}, their output taxonomies are evaluated whole without partitioning.
In addition, the two metrics for novel topic discovery (i.e., relation accuracy and subtopic integrity) are designed to evaluate the topic taxonomy expansion methods, so it is infeasible to measure the aspects on the output taxonomies of \hlda and \taxogen.
Thus, we only report the metric for topic identification (i.e., term coherence) in Table~\ref{tbl:humaneval}.

\smallsection{Term Coherence}
It indicates how strongly terms in a topic node are relevant to each other. 
Evaluators count the number of terms that are relevant to the common topic (or topic name) among the top-5 terms found for each topic node.

\smallsection{Relation Accuracy}
It computes how accurately a topic node is inserted into a given topic hierarchy (i.e., \textit{precision} for novel topic discovery).
For each valid position, evaluators count the number of newly-inserted topics that are in the correct relationship with the surrounding topics.

\smallsection{Subtopic Integrity} 
It measures the completeness of subtopics for each topic node (i.e., \textit{recall} for novel topic discovery).
Evaluators investigate how many ground-truth novel topics, which were deleted from the original taxonomy, match with one of the newly-inserted topics.

\subsection{Examples of Topic Phrase Generation}
\label{subsec:examples}
\begin{figure}[t]
\centering
\begin{subfigure}{\linewidth}
    \centering
    \includegraphics[width=\linewidth]{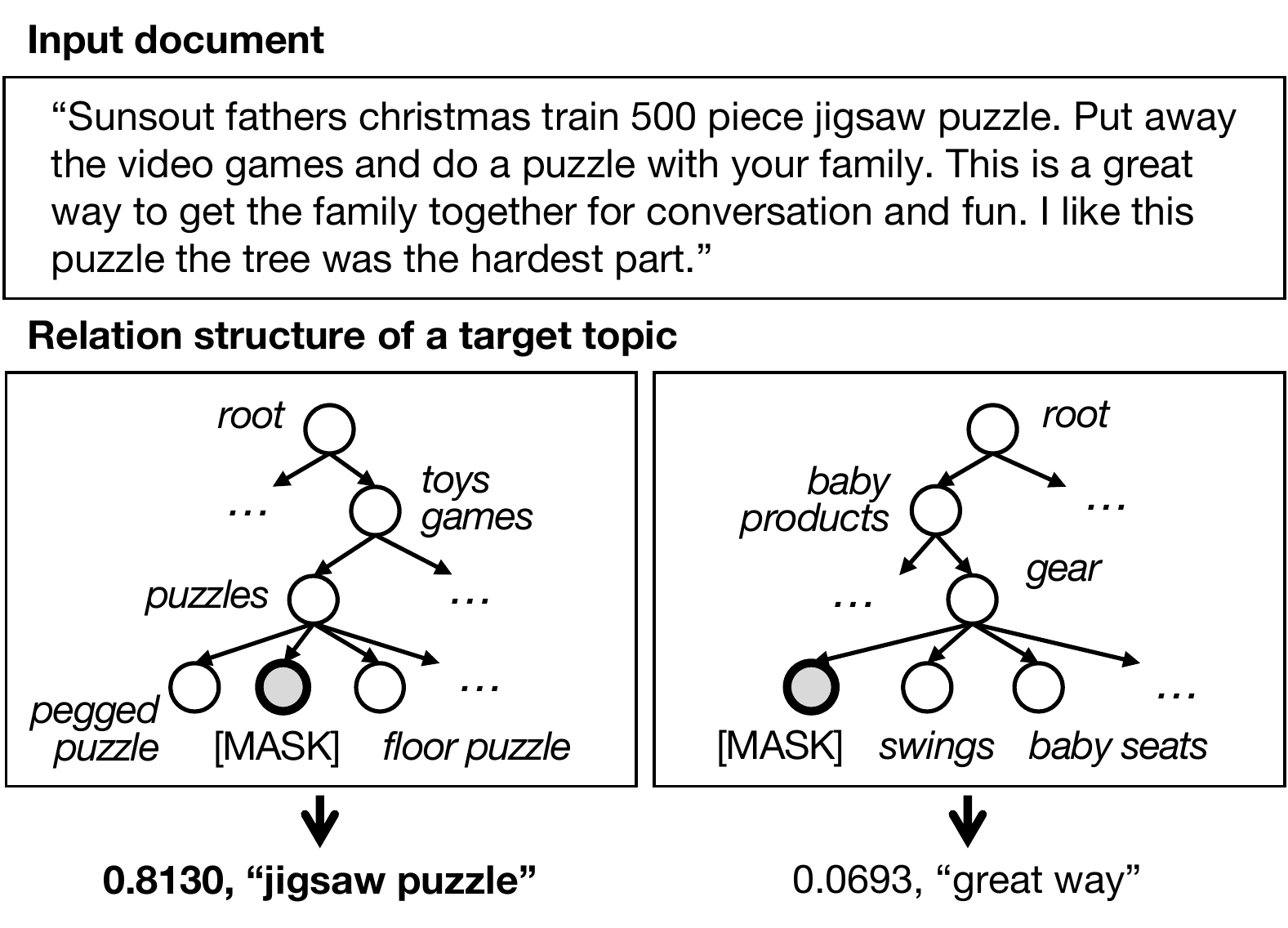}  
    \caption{Dataset: \amazon}
\end{subfigure}
\begin{subfigure}{\linewidth}
    \centering
    \includegraphics[width=\linewidth]{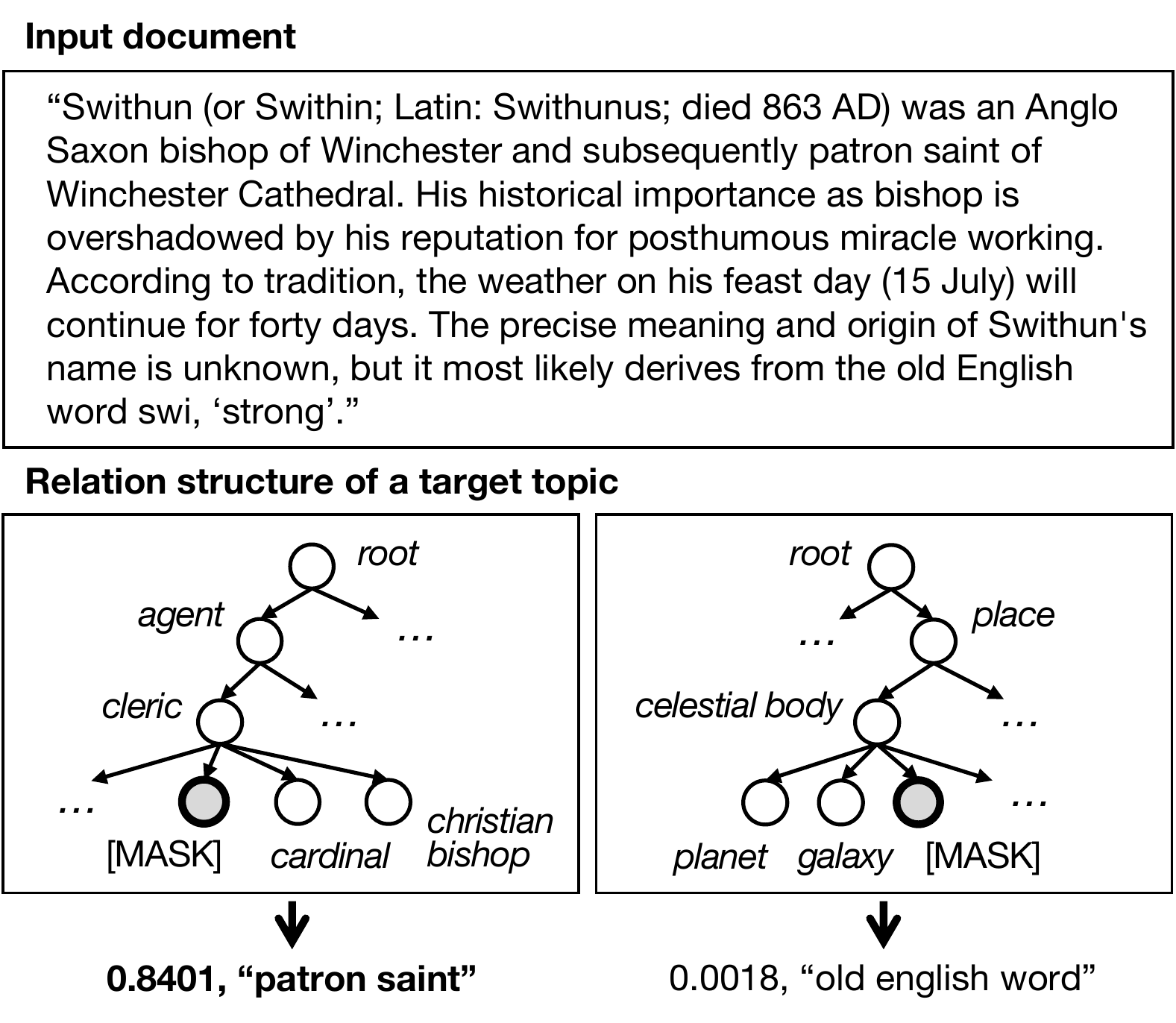}
    \caption{Dataset: \dbpedia}
\end{subfigure}
\caption{Examples of topic-conditional phrase generation, given a document and its relevant/irrelevant topic.}
\label{fig:examples}
\end{figure}

We provide additional examples of topic-conditional phrase generation, obtained by \proposed.
Figure~\ref{fig:examples} illustrates a confident phrase (Left) and a non-confident phrase (Right), generated from each input document and the given relation structure of a target topic, for both datasets.
As discussed in Section~\ref{subsubsec:casestudy}, 
in case that a target topic is relevant to the document (i.e., high topic-document similarity score), \proposed successfully generates a phrase relevant to the target topic.
On the other hand, in case that a target topic is irrelevant to the document (i.e., low topic-document similarity score), \proposed obtains a phrase irrelevant to the target topic.

\end{document}